\documentclass[journal]{IEEEtran}
   \usepackage[pdftex]{graphicx}
\usepackage{multicol, blindtext}
\usepackage{amsmath}
\usepackage{enumerate}
\begin{document}
\title{Detection of Anomalous Crowd Behavior Using Spatio-Temporal Multiresolution Model and Kronecker Sum Decompositions}


\author{Kristjan~Greenewald,~\IEEEmembership{Student Member,~IEEE,}
        and~Alfred O.~Hero III,~\IEEEmembership{Fellow,~IEEE}

\thanks{K. Greenewald and A. Hero III are with the Department
of Electrical Engineering and Computer Science, University of Michigan, Ann Arbor,
MI, USA. This document has been cleared for public release: PA Approval Number: 88ABW-2013-4708. }
}
\maketitle
\begin{abstract}
In this work we consider the problem of detecting anomalous spatio-temporal behavior in videos. Our approach is to learn the normative multiframe pixel joint distribution and detect deviations from it using a likelihood based approach. Due to the extreme lack of available training samples relative to the dimension of the distribution, we use a mean and covariance approach and consider methods of learning the spatio-temporal covariance in the low-sample regime. Our approach is to estimate the covariance using parameter reduction and sparse models. The first method considered is the representation of the covariance as a sum of Kronecker products as in \cite{greenewaldArxiv}, which is found to be an accurate approximation in this setting. We propose learning algorithms relevant to our problem. We then consider the sparse multiresolution model of \cite{Choi2010gaussian} and apply the Kronecker product methods to it for further parameter reduction, as well as introducing modifications for enhanced efficiency and greater applicability to spatio-temporal covariance matrices. We apply our methods to the detection of crowd behavior anomalies in the University of Minnesota crowd anomaly dataset \cite{UMNData}, and achieve competitive results.
\end{abstract}

\section{Introduction}

The detection of changes and anomalies in imagery and video is a fundamental problem in machine vision. Traditional change detection has focused on finding the differences between static images of a scene, usually observed in a pair of images separated in time (frequently hours or days apart). As video surveillance sensors have proliferated, however, it has become possible to analyze the spatio-temporal characteristics of the scene. The analysis of the behavior of crowds (particularly of humans and/or vehicles) has become essential, with the detection of both abnormal crowd motion patterns and individual behaviors being important surveillance applications. In this paper, we focus on the problem of modeling and detecting changes and anomalies in the spatio-temporal characteristics of videos. As an application, we consider the detection of anomalous crowd behaviors in several video datasets. Our methods provide pixel-based spatio-temporal models that enable the detection of anomalies on any scale from individuals to the crowd as a whole.


\subsection{Approach}

The two main approaches to crowd video anomaly detection are those based on first extracting the tracks and finding anomalous track configurations (microscopic methods) and those that are based directly on the video without extracting the tracks (macroscopic) \cite{wu2010chaotic}. An example of the microscopic approach is the commonly used social force model of \cite{helbing1995social}. The macroscopic methods tend to be the most attractive in dense crowds because track extraction can be computationally intensive in a crowd setting, and in dense crowds track association becomes extremely difficult and error-prone \cite{wu2010chaotic}. In addition, it is frequently the case that long-term individual tracks are irrelevant because the characteristics of the crowd itself are of greater interest. We thus follow the macroscopic approach.

In this work, we focus on learning generative joint models of the pixels of the video themselves as opposed to the common approach of feature extraction. Our reasons for this include improving the expressivity of the model, improving the general applicability of our methods, minimizing the assumptions that must be made about the data, and eliminating preprocessing and the associated loss of information.

\begin{figure}[htb]
\centering
\includegraphics[width=2.85in]{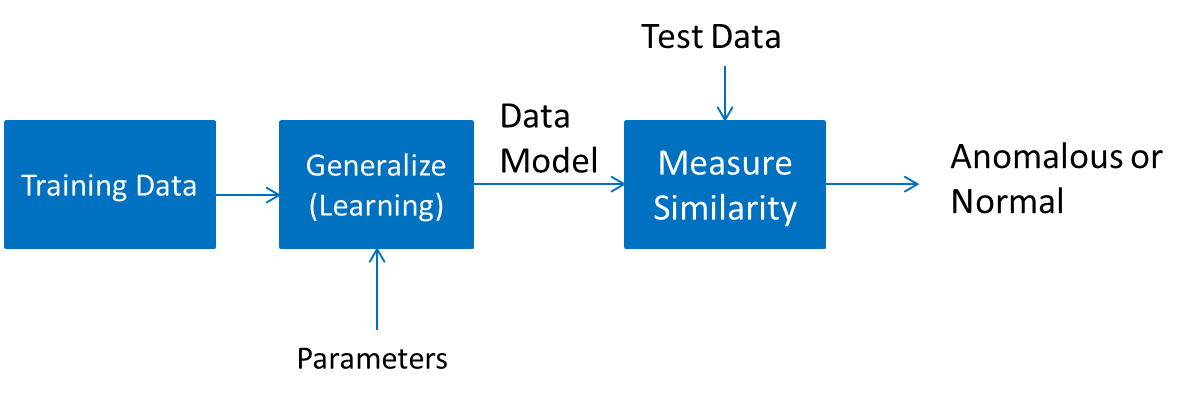}
\caption{Anomaly detection block diagram}
\label{Fig:FlowChart}
\end{figure}

We follow a statistical approach to anomaly detection \cite{chandola2009anomaly}, that is, we learn the joint distribution of the nonanomalous data and declare data that is not well explained by it in some sense to be anomalies. A block diagram of the process is shown in Figure \ref{Fig:FlowChart}.

We thus need to learn the joint distribution of the video pixels. In order to learn the temporal as well as the spatial characteristics of the data, the joint distribution of the pixels across multiple adjacent video frames must be found. For learning, we make the usual assumption that the distribution is stationary over the learning interval. To make the assumption valid, it may be necessary to limit the length of the learning interval and hence the number of samples. Our approach is to learn the distribution for a finite frame chunk size $T$, that is, the $T$ frame $N$ spatio-temporal pixel joint distribution $p(X = \{x_n \}_{n=t-T}^{t-1} )$. Once this distribution is learned, it can be efficiently extended to larger frame chunk sizes according to either an AR (Markov), MA, or ARMA process model \cite{wiesel2013time}. Limiting $T$ reduces the number of learned parameters and thus the order of the process, hence reducing the learning variance.

In order to reduce the number of samples required for learning, we use the parametric approach of learning only the mean and covariance of $X$. The number of samples required by standard covariance methods to achieve low estimation variance grows as $O(NT \log NT )$. Hence, the spatio-temporal ``patch size" of the learned distribution that these methods can handle is still severely limited. Making the ``patch size" as large as possible is highly desirable as it allows for the modeling of interactions (such as between different regions of a crowd) across much larger spatial and temporal intervals, thus allowing for larger-scale relational type anomalies to be detected.

The number of samples required for covariance learning can be vastly reduced if prior assumptions are made, usually in the form of structure imposition and/or sparsity. In this paper, we use the sums of Kronecker products covariance representation, as well as a sparse tree-based multiresolution model which incorporates both structural and learned sparsity and has a degree of sparsity that is highly tunable. The multiresolution approach is particularly attractive because it defines hidden variables that allow analysis of the video at many different scales.

Once we have the estimate of the spatio-temporal pixel distribution, the typical statistical approach \cite{chandola2009anomaly} is to determine whether or not a video clip is anomalous by evaluating its likelihood under the learned distribution. This approach is based on the fact that the anomalous distribution is unknown, so likelihood ratio tests are inappropriate. As a measure of how well the model explains the data, the likelihood is an attractive feature. Thresholding the likelihood is supported by theoretical considerations \cite{chandola2009anomaly}.

In this work, we use the Mahanolobis distance
\begin{equation}
\label{Eq:Mahanolobis}
\ell(x) = (x-\mu_x)^T \mathbf{\Sigma}^{-1} (x-\mu_x)
\end{equation}
which is the negative loglikelihood under the multivariate Gaussian assumption. Although the data is not exactly multivariate Gaussian, the metric is convenient and robust and its use in non-Gaussian problems is well supported. In addition, it allows for evaluation of the likelihood of local subregions by extracting submatrices from $\Sigma^{-1}$ (conditional distribution) or $\Sigma$ (marginal distribution), and hence allowing for relatively efficient localization of the anomalous activity. We discuss the details of likelihood thresholding in Section \ref{S:Anom}.

\subsection{Previous Work on Detection of Crowd Anomalies}
In this section, we briefly review select previous work on the detection of crowd anomalies using non track based (macroscopic) approaches.

Many macroscopic techniques are based on computing the optical flow in the video, which attempts to estimate the direction and magnitude of the flow in the video at each time instant and point in the scene. In \cite{chen2011motion} optical flow is used to cluster the movement in the scene into groups (hopefully of people), and models inter group interactions using a force model. Anomalies are declared when the observed ``force" is anomalously large or unexpected. Particle advection, which is based on optical flow, has also been used. The authors of \cite{mehran2009abnormal} compute the optical flow of the video and use it to advect sets of particles. Social force modeling is then performed on the particles, and anomalies are declared based on a bag of words model of the social force fields. \cite{wu2010chaotic} computes chaotic invariants on the particle advection trajectories. The chaotic invariants are then modeled using a Gaussian mixture model for anomaly declaration.

Various spatio-temporal features have also been used. In \cite{mahadevan2010anomaly}, the authors model the video by learning mixtures of dynamic textures, that is, patchwise multivariate state space models for the pixels. This is slightly similar to our approach in that it models the pixels directly using a Gaussian model. The patch size they use, however, is severely limited (\cite{mahadevan2010anomaly} uses $13\times13$ blocks) due to the sample paucity issues which are our main focus in this work. A spatio-temporal feature-based blockwise approach using K nearest neighbors for anomaly detection is given in \cite{saligrama2012video}, and \cite{benezeth2009abnormal} uses a cooccurence model. A gradient based approach is used in \cite{luvison122011automatic}. They divide the video into cuboids, compute the spatiotemporal gradients, and model them using sparse KDE to get likelihoods and declare anomalies accordingly.

\subsection{Outline}
The outline of the remainder of the paper is as follows. In Section \ref{S:Kron} we discuss the sums of Kronecker products covariance representation and its application to video, as well as introduce a new estimation algorithm. In Section \ref{S:MultiRes} we review the sparse multiresolution model of Choi et al. In Section \ref{S:Mod} we present our modifications of and application to spatio-temporal data. Our approach to anomaly detection using the learned model is presented in Section \ref{S:Anom}. Section \ref{S:Results} presents video anomaly detection results, and Section \ref{Conclusion} concludes the paper.

\section{Kronecker Product Representation of Multiframe Video Covariance}
\label{S:Kron}
In this section, we consider the estimation of $\Sigma$ using sums of Kronecker products. Additional details of this method are found in our paper \cite{greenewaldArxiv} and in \cite{tsiliArxiv}.
\subsection{Basic Method}

As the size $NT$ of $\Sigma$ can be very large, even for moderately large $N$ and $T$ the number of degrees of freedom ($NT(NT+1)/2$) in the covariance matrix can greatly exceed the number $n$ of i.i.d. samples available to estimate the covariance matrix. One way to handle this problem is to introduce structure and/or sparsity into the covariance matrix, thus reducing the number of parameters to be estimated. In many spatio-temporal applications it is expected (and confirmed by experiment) that significant sparsity exists in the inverse pixel correlation matrix due to Markovian relations between neighboring pixels and frames. Sparsity alone, however, is not sufficient, and applying standard sparse methods such as GLasso directly to the spatio-temporal covariance matrix is computationally prohibitive \cite{tsiligkaridis2013convergence}.

\begin{figure}[htb]
\centering
\includegraphics[width=2.85in]{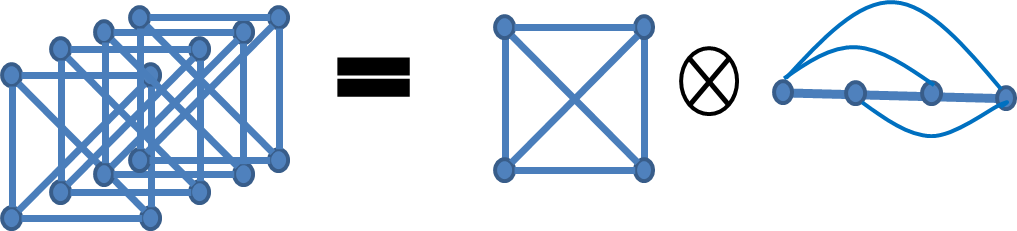}
\caption{Gaussian graphical model representation of Kronecker product decomposition}
\label{Fig:GraphKron}
\end{figure}

A natural non-sparse alternative is to introduce structure is by modeling the covariance matrix $\mathbf{\Sigma}$ as the Kronecker product of two smaller matrices, i.e.
\begin{equation}
\label{KronApprox}
\mathbf{\Sigma} = \mathbf{T}\otimes \mathbf{S}.
\end{equation}
thus reducing the number of parameters from $pq(pq+1)/2$ to $p(p+1)/2+q(q+1)/2$ where ($\Sigma(pq\times pq), S(p\times p),T(q\times q)$). The equivalent graphical model decomposition is shown in Figure \ref{Fig:GraphKron}. When  the measurements are Gaussian with covariance of this form they are said to follow a matrix-normal distribution \cite{tsiligkaridis2013convergence}. This model lends itself to coordinate decompositions \cite{tibshirani,tsiliArxiv,greenewaldArxiv}. For spatio-temporal data, we consider the natural decomposition of space (pixels) vs. time (frames) as done in \cite{greenewaldArxiv}. In this setting, the $\mathbf{S}$ matrix is the ``spatial covariance" and $\mathbf{T}$ is the ``time covariance."

Previous applications of the model of Equation \eqref{KronApprox} include MIMO wireless channel modeling as a transmit vs. receive decomposition \cite{werner2007estimation}, geostatistics \cite{Cressie1993}, genomics \cite{yin2012model}, multi-task learning \cite{bonilla2008multi}, collaborative filtering \cite{yu2009large}, face recognition \cite{zhang2010learning}, mine detection \cite{zhang2010learning}, recommendation systems \cite{tibshirani}, wind speed prediction \cite{genton2007separable} and prediction of video features \cite{greenewaldArxiv}.

An extension to the representation \eqref{KronApprox} introduced in \cite{tsiliArxiv} approximates the covariance matrix using a sum of Kronecker product factors
\begin{equation}
\label{SumApprox}
\mathbf{\Sigma} \approx \sum_{i=1}^{r} \mathbf{T}_i \otimes \mathbf{S}_i
\end{equation}
where $r$ is the separation rank.

\begin{figure}[htb]
\centering
\includegraphics[width=2.85in]{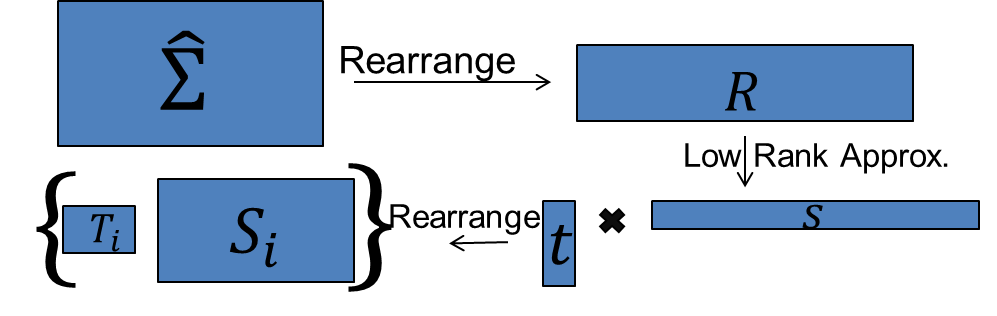}
\caption{Basic LS sum of Kronecker products approximation algorithm.}
\label{Fig:LSAlg}
\end{figure}

This allows for more accurate approximation of the covariance when it is not in Kronecker product form but most of its energy is in the first few Kronecker components. An algorithm for fitting the model \eqref{SumApprox} to a measured sample covariance matrix was introduced in \cite{tsiliArxiv} called Permuted Rank Least Squares (PRLS) and was shown to have strong high dimensional guarantees in MSE performance. The LS algorithm is an extension of the Kronecker product estimation method of \cite{werner2008estimation} and is based on the SVD. In \cite{tsiliArxiv}, some regularization is also introduced. In this work, we use different regularization methods. A pictorial representation of the basic LS algorithm is shown in Figure \ref{Fig:LSAlg}.

%

\subsection{Diagonally Corrected Method}
In \cite{greenewaldArxiv}, we presented a method of performing sum of Kronecker products estimation with a modified LS objective function that ignores the errors on the diagonal. The diagonal elements are then chosen to fit the sample variances, with care to choose them so as to guarantee positive semidefiniteness of the the overall estimate. The main motivation for this method is that uncorrelated variable noise occurs in most real systems but damages the Kronecker structure of the spatio-temporal covariance. This also allows for gain in expressivity when doing diagonal regression \cite{greenewaldArxiv}. The weighted LS solution is given by the alternating projections method (iterative).

\subsection{Temporal Stationarity}
\label{S:Toeplitz}
Since we are modeling a temporal process with a length much longer than $T$, the spatio-temporal covariance that we learn should be stationary in time, that is, $\Sigma$ should be block Toeplitz \cite{wiesel2013time}. If all the $T_i$ matrices are Toeplitz, then $\Sigma$ is block Toeplitz. Furthermore, if $\Sigma$ is block Toeplitz and has separation rank $r$, a Kronecker expansion exists where every $T_i$ is Toeplitz. We thus only need to estimate the value of the diagonal and each superdiagonal of the $T_i$ ($2T-1$ parameters). To estimate these parameters, we use the method of \cite{kamm2000optimal}, which uses the same LS objective function as in \cite{werner2008estimation, tsiliArxiv, greenewaldArxiv} with the additional Toeplitz constraint. The equivalent (rearranged) optimization problem is given by
\begin{align}
\min_{\hat{R}, rank(\hat{R}) = r} || R - \hat{R} ||_F^2\\\nonumber
\hat{R} = \sum_{i = 1}^r t_i s_i^T\\\nonumber
s.t. \: T_i\: \mathrm{Toeplitz} \quad \forall i
\end{align}

The Toeplitz requirement is thus equivalent to
\begin{equation}
\label{Eq:Toeplitz}
[t_i]_{k} = u^{(i)}_{j + T},\: \forall k \in \mathcal{K}(j), j \in [-T+1,\: T-1]
\end{equation}
for some vector $u^{(i)}$ where
\begin{equation}
\mathcal{K}(j) = \{k : \: (k-1)T + k + j \in [-T+1,\: T-1]\}
\end{equation}
Clearly $| \mathcal{K}(j) | = T - |j|$. Let
\begin{equation}
\label{Eq:Tweight}
\tilde{t}^{(i)}_{j+T}= u^{(i)}_{j + T}\sqrt{T - |j|}, \quad j \in [-T+1,\: T-1]
\end{equation}
for reasons that will become apparent.

We now formulate this problem as a LS low rank approximation problem. We use the notation $R_k$ to denote the $k$th row of $R$. Let $\tilde{R}$ be given by
\begin{equation}
\tilde{R} =\sum_{i = 1}^r \tilde{t}^{(i)} s_i^T
\end{equation}
Then
\begin{align}
\label{Eq:ToeSol}
&\arg \min_{\hat{R}} || R - \hat{R} ||_F^2 \\\nonumber
&=\arg \min_{\hat{R}}  \sum_{j = 1-T}^{T-1} \sum_{k \in \mathcal{K}(j)} \left\| R_{k} - \frac{1}{\sqrt{T - |j|}}\tilde{R}_{j+T}\right\|_F^2\\\nonumber
&= \arg \min_{\hat{R}}  \sum_{j = 1-T}^{T-1} \sum_{k \in \mathcal{K}(j)} - 2R_k \frac{1}{\sqrt{T - |j|}}\tilde{R}_{j+T}^T + \frac{1}{T - |j|}\tilde{R}_{j+T}\tilde{R}_{j+T}^T\\\nonumber
&= \arg \min_{\hat{R}} \sum_{j = 1-T}^{T-1} - 2\left(\frac{1}{\sqrt{T - |j|}}\sum_{k \in \mathcal{K}(j)}R_k\right) \tilde{R}_{j+T}^T + \tilde{R}_{j+T}\tilde{R}_{j+T}^T\\\nonumber
&= \arg \min_{\hat{R}}  \sum_{j = 1-T}^{T-1} \left\| \left(\frac{1}{\sqrt{T - |j|}}\sum_{k \in \mathcal{K}(j)}R_k\right) - \tilde{R}_{j+T}\right\|_F^2\\\nonumber
&= \arg \min_{\tilde{R}}  \left\| B - \tilde{R}\right\|_F^2
\end{align}
where
\begin{equation}
B_{j+T} = \frac{1}{\sqrt{T - |j|}}\sum_{k \in \mathcal{K}(j)}R_k  \quad \forall j \in [-T+1, \: T-1]
\end{equation}
and the low rank constraint is clearly manifested as
\begin{equation}
rank(\tilde{R})  \leq r
\end{equation}

This is now in the desired low-rank form and thus solvable using the SVD or by one of the other weighted LS methods in this section.

The matrices $T_i$ are found from the $\tilde{t}^{(i)}$ (left singular vectors) by unweighting according to \eqref{Eq:Tweight} and expanding them into the $t_i$ using \eqref{Eq:Toeplitz} and then rearranging.

The use of this method in the context of the PRLS regularization is clear. An additional benefit is that the size of the low rank problem has been decreased by a factor of $(2T-1)/T^2$.

Using this method, it is clear that any block Toeplitz matrix $\Sigma$ can be expressed without error using a sum of $2T-1$ Kronecker products. Due to symmetry, however, the matrix can be completely determined using $T$ Kronecker products.

Due to the structure of the result \eqref{Eq:ToeSol} being the same as the original optimization problem (i.e. low rank approximation, where the right singular vectors are still $s_i$), Toeplitz structure can be enforced in both the $T_i$ and $S_i$ by starting with the result of \eqref{Eq:ToeSol} and repeating the derivations in \eqref{Eq:ToeSol} except with the $s_i$ constrained.

When it is desirable to learn an infinite-length AR, MA, or ARMA model by finding the $k$ banded block Toeplitz covariance as in \cite{wiesel2013time}, this method is particularly attractive because in order to do the LS estimate, it is only necessary to find the $k$ frame sample covariance and do Toeplitz approximation with the $\sqrt{T-|j|}$ weights removed (i.e. $T\rightarrow \infty$).

\subsection{Sum of Kronecker Products for Nonrectangular Grids}
\label{S:Flow}

While with enough terms any covariance can be represented as a sum of Kronecker products, the separation rank is significantly lower for those matrices that are similar to a single Kronecker product. When ``flow" is occuring through the variables in time, or equivalently the best tree defined on the spatiotemporal data is nonrectangular across frames, variables (pixels) of the same index do not correspond across frames. This results in a flow of correlations through the variables as the time interval increases. This situation is produces a covariance matrix that has a very non-Kronecker structure. If, however, we shift the indexes of the variables in each frame so that corresponding (highly correlated, or adjacent in a tree graph) pixels have the same index, the approximate Kronecker structure usually returns or at least improves. The mapping is usually not one-to-one, however, so some variables in a frame will not have corresponding variables in the next frame.

\begin{figure}[htb]
\centering
\includegraphics[width=2.85in]{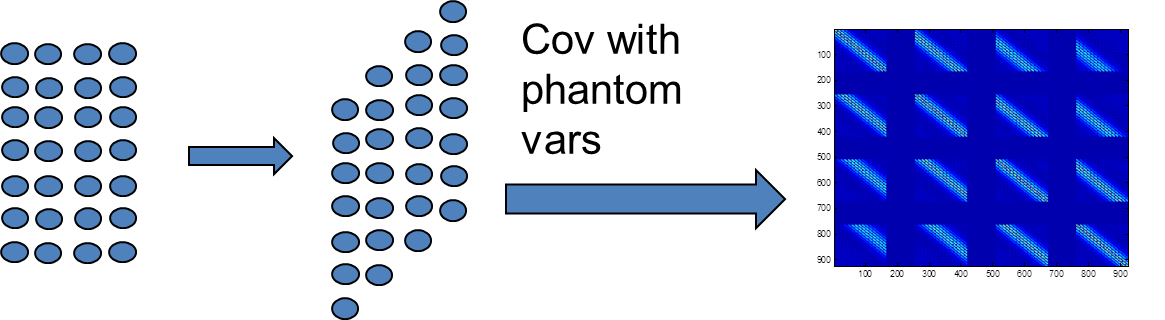}
\caption{Kronecker approximation method for nonrectangular variable grids. The nonrectangular grid is embedded in a rectangular grid with dummy variable padding. The Kronecker product representation of the new rectangular grid is then found using weighted least squares.}
\label{Fig:NonRect}
\end{figure}

To handle this, we set up a larger ($(N+(T-1)\Delta N) \times T$) space-time rectangular grid of variables that contains within it the nonrectangular ($N\times T$) grid of pixel variables defined by the required index shifting, and has the remaining variables be dummy variables. This unified grid of variables is then indexed according to the rectangular grid. See the two leftmost images in Figure \ref{Fig:NonRect}. The covariance matrix of the complete set of variables then has valid regions corresponding to the real variables, and dummy regions corresponding to the dummy variables. As an example, see the last pane in Figure \ref{Fig:NonRect}, where the dark regions correspond to the dummy regions. If the dummy regions are allowed to take on any values we choose, it is clear that there is indeed much better Kronecker structure using the Kronecker dimensions $T$ and $N+(T-1)\Delta N$. To handle this for Kronecker approximation, we merely remove (don't penalize) the terms of the standard LS objective (approximation error) function corresponding to the dummy regions of the covariance, thus allowing them in a sense to take on the values most conducive for low separation rank. After the rearrangement operator in the LS algorithm, the problem is a low-rank approximation problem with portions of the matrix to be approximated having zero error weight. This is the standard weighted LS problem, which can be solved iteratively as mentioned above. Finally, after the approximating covariance is found, the valid regions are extracted and reindexed to obtain the covariance of the nonrectangular grid.

\subsection{Examples of Low Separation Rank Processes}
An example of multiple Kronecker structure is the traveling wave field
\begin{align}
&h(x,y)\sin(g(x,y) - ct)\\\nonumber
& = h(x,y)\sin(g(x,y))\cos(ct) - h(x,y)\cos(g(x,y))\sin(ct)
\end{align}
Since the Kronecker representation is exact for separable processes, two Kronecker components are required to perfectly capture the covariance of a spatio-temporal wave, although one will capture necessary information such as wavelength, nature of $g(x,y)$, speed, and amplitude.

\section{Sparse Multiresolution Model}
\label{S:MultiRes}

While the sum of Kronecker products representation reduces the number of parameters considerably, most images are too large to be able to estimate or even form (due to memory issues) the $S_i$ matrices directly. In addition, since video characteristics do vary over space, the Kronecker decomposition can break down as the spatial patch size increases. Hence further parameter reduction is needed. Simple approaches include considering block diagonal covariance estimation, where the video is divided into spatial blocks for estimation, and/or by enforcing spatial stationarity as done in the temporal dimension. Additional reduction can be achieved by using sums of triple Kronecker products which forces slowly changing characteristics over sets of blocks using windowing. An issue, however, with doing these blockwise decompositions is that correlations between neighboring pixels in different blocks are ignored.

\begin{figure}[htb]
\centering
\includegraphics[width=2.85in]{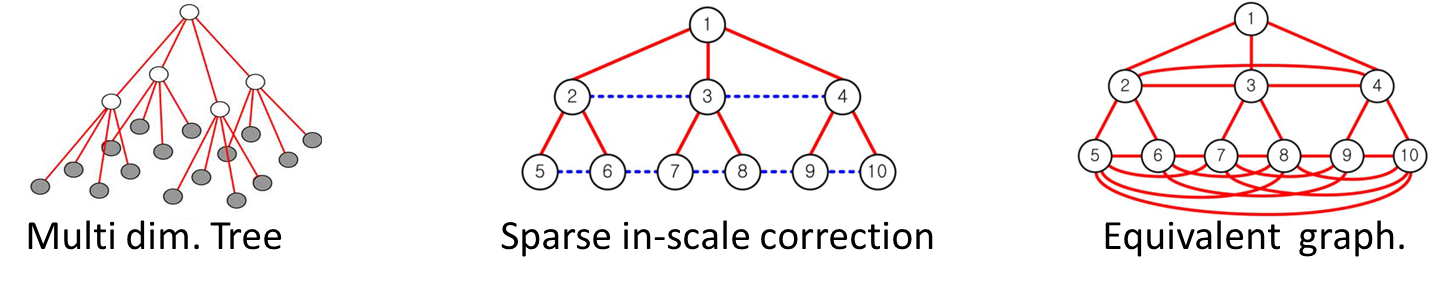}
\caption{Multiresolution models: Tree, Tree with sparse in-scale covariance (conjugate graph), Equivalent graphical model}
\label{Fig:SIM}
\end{figure}

We thus consider the use of tree based multiresolution models. As a starting point, we consider the multiresolution model of Choi et al \cite{Choi2010gaussian} and modify it for our problem. Choi et al's sparse covariance model, which they refer to as a sparse in-scale conditional covariance multiresolution (SIM) model, starts with a Gaussian tree with the observed variables on the bottom row and adds sparse in-scale covariances (conditioned on the other levels) to each level (see Figure \ref{Fig:SIM}). The added in-scale covariances are introduced because Gaussian trees are not expressive enough and introduce artifacts such as blocks.


\subsection{Trees}
In order to clarify the next section, we briefly review the basic Gaussian tree model. The model is a Gaussian graphical model with the variables $x(i)$ and connections arranged in a tree shape, that is, every variable $x(i)$ is either the root node or has a single parent $x(p(i))$, and can have multiple children. The edge and node parameters are usually expressed implicitly by viewing the tree as a Markov chain beginning at the root node. That is, each child variable is given by \cite{kannan2000ml}
\begin{align}
x(i) = a(i)x(p(i)) + n_i\\\nonumber
n_i \sim \mathcal{N}(0,Q(i))
\end{align}
The extension to multivariate nodes is simple \cite{kannan2000ml}.

\subsection{Inference}
In order to use this model (especially for evaluating likelihoods), it is necessary to be able to infer the hidden variables given observed variables. For our application, we observe the bottom level variables and infer the upper levels.

The general formulation is that we observe a linear combination $y$ of a set of the variables $(y=Cx)$ plus noise with covariance $R$, and infer the variables $x$ via maximum likelihood estimation under a multivariate Gaussian model. In what follows, we refer to the information matrix associated with the tree with the diagonal elements removed as $J^h$ and the added in-scale conditional covariance matrices arranged blockdiagonally as $\Sigma^c$. $\Sigma^c$ is a blockwise positive semidefinite matrix with nonzero values only between variables in the same level. As a result, the overall information matrix of the multiresolution model is $J^h + (\Sigma^c)^{-1}$ since the inversion of a blockdiagonal matrix is blockdiagonal with the blocks being the inverses of the original blocks.

The MLE solution is \cite{Choi2010gaussian} is to solve for $x$ in
\begin{equation}
(J^h+(\Sigma^c )^{-1}+J^p ) x = h
\end{equation}
where
\begin{align}
h = C^T R^{-1} y\\\nonumber
J^p =C^T R^{-1}C
\end{align}

\begin{figure}[htb]
\centering
\includegraphics[width=1.85in]{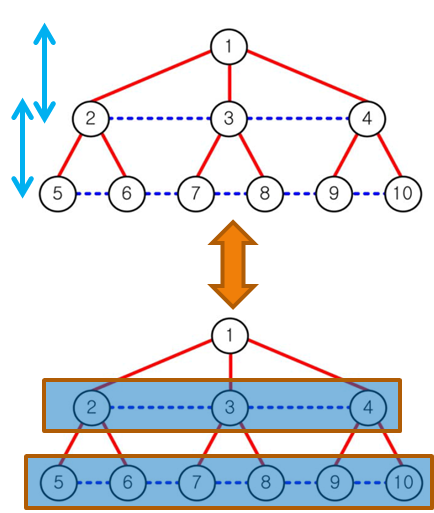}
\caption{Representation of multiresolution inference model of \cite{Choi2010gaussian}. Alternate between between scale and in scale inference.}
\label{Fig:Inference}
\end{figure}

The approach of \cite{Choi2010gaussian} is to exploit the sparsity of the tree model and the in-scale covariance corrections via matrix splitting. In particular, the solution is found iteratively by alternating between solving (Figure \ref{Fig:Inference})
\begin{equation}
(J^h + J^p + D)x_{new} = h - \Sigma_c^{-1}x_{old} + Dx_{old}
\end{equation}
for $x_{new}$ (between scale inference) using an appropriate iterative algorithm and computing the sparse matrix vector multiplication (in-scale inference)
\begin{equation}
x_{new} = \Sigma_c(h-(J^h+J^p) x_{old}).
\end{equation}
The term $\Sigma_c^{-1} x_{old}$ can be computed by solving the sparse system of equations $\Sigma_c z = x_{old}$ for $z$. Hence each iteration is performed in approximately linear time relative to the nonzero elements in the sparse matrices.


In the case where we observe a portion of the variables ($C = [I_{n_1} 0_{n-n_1}]$) without noise ($J^p = \infty$) (as in our application), it is straightforward to modify the above equations using the standard conditional MLE approach as in \cite{yu2012copula}, resulting in greatly enhanced computational complexity and numerical precision. 


\subsection{Learning}
The learning algorithm proposed in \cite{Choi2010gaussian} is based on learning the tree first and then correcting it with the in-scale covariances. To learn the tree, first specify (or learn) a tree structure where the bottom layer is the observed variables. Once the tree structure is specified, the parameters (edge weights and node noise variances) are learned from the spatio-temporal training samples using the standard EM algorithm \cite{kannan2000ml}. We use the tree to represent the covariance only, thus the training data has the mean subtracted to make it zero mean.

The in-scale covariances are then learned to eliminate the artifacts that arise in tree models. The first step of the approach is to pick a target covariance of the observed variables (the sample covariance in \cite{Choi2010gaussian}) and determine the target in-scale conditional information matrices that would result in the target covariance being achieved. The target in-scale conditional information matrices are found using a recursive bottom up approach
\begin{equation}
\label{Eq:BUP}
\Sigma_{[m]} = A_m \Sigma_{[m+1]} A_m^T + Q_m
\end{equation}
where $A_m$ and $Q_m$ are determined by $J_{tree}$ \cite{Choi2010gaussian}.

The target information matrix can be computed to be \cite{Choi2010gaussian}
\begin{equation}
\label{Eq:LearnInv}
J_{[m]}^* = \Sigma_{[m]}^{-1} + J^*_{[m],c} (J^*_c)^{-1}J_{c,[m]}^* +J^*_{[m],f} (J_f^*)^{-1} J_{f,[m]}^*
\end{equation}
by setting the marginal covariance equal to the target covariance.

Regarding computational complexity, it is important to note that the method requires the inversion of the target covariances $\Sigma_{[m]}$ at each level, thus making the learning complexity at least $O((NT)^3)$. This can be a severe bottleneck for our application. We propose a method of dramatically reducing this cost in the next section.

Secondly, the target in-scale conditional covariances are sparsified. It is well known that applying GLASSO style logdet optimization (regularization) \cite{Choi2010gaussian} to a matrix sparsifies its inverse while maintaining positive semidefiniteness. Hence, applying the method to the target information matrices results in sparse target covariances \cite{Choi2010gaussian}. This gives the sparse conditional in-scale covariances as required for the model. Methods of determining all the in-scale covariances jointly are also presented in \cite{Choi2010gaussian}, but we do not consider them here due to their computational complexity.


\subsection{Modifications for Space-Time Data}
\label{S:Mod}

In this section, we develop appropriate modifications of the multiresolution model of the previous section in order to apply it to spatio-temporal data.


%
%
\subsubsection{Structure: Spatial Tree Only}
Complete stationarity in the tree model itself can be achieved by decoupling the frames (with the interframe connections filled in later using the in-scale covariance corrections). This is equivalent to having the tree model the spatial covariance only. Hence there are substantial computational and parameter reduction benefits to this approach. Another option which we do not employ here is to use space-based priors when learning the tree.

\subsubsection{Subtree Based Learning}
In many of the applications we consider, it is desirable to estimate the multiframe covariance more than once (e.g. to compare different portions of frame sequences). Hence, the $O(N^3T^3)$ complexity of the inversion required to compute the target information matrix at the lower scales is prohibitive for video data. Hence, we propose to only consider local in-scale connections. 

Our approach is to force the inscale conditional information matrices $J_{[m]}^*$ to be blockdiagonal, at least for the lower levels. As a result, Equation \label{Eq:LearnInv} indicates that only the corresponding block diagonal elements of the right hand side are required, giving substantial computational savings immediately. Additional savings are achieved by using a local estimate for the blocks of $\mathbf{\Sigma}^{-1}$, i.e. estimating a block of $\mathbf{\Sigma}$ containing but somewhat larger than the block of interest, inverting, and extracting the relevant portion. This is based on the notion that the interactions relevant to local conditional dependencies should also be local, and makes the algorithm much more scalable. This is a particularly good approximation when the dominant interactions are local.

To get the upper level target covariances, it is not necessary to form the bottom level sample covariance, as, following the recursion of \eqref{Eq:BUP},
\begin{align}
\label{Eq:BUPEff}
\Sigma_{[m]} &= \frac{1}{N_S}(A^{(m)} X)(A^{(m)} X)^T\\\nonumber
 &+ \sum\nolimits_{m' = m}^{M} \left(\prod\nolimits_{k=m}^{m'-1} A_k \right)Q_{m'}\left(\prod\nolimits_{k=m}^{m'-1} A_k \right)^T
\end{align}
where $N_S$ is the number of samples, $X$ is the matrix of samples, $m$ is the current level, and $A^{(m)} = \prod\nolimits_{m'=m}^{M} A_{m'}$. If necessary, regularization etc. is applied to the first term of \eqref{Eq:BUPEff}. This allows for interblock connections at as low a level as possible to minimize blockwise artifacts.

\subsubsection{Kronecker}

In the original multiresolution model, parameter reduction for the in-scale covariances is achieved using sparsity. Naturally, we wish to use the Kronecker PCA representation for the covariance to reduce the number of parameters. We use DC-KronPCA on the first term of \eqref{Eq:BUPEff}. This is possible because the tree is in the spatial dimension only, hence the multiplication with $\mathbf{A}$ matrices to move through the levels does not affect the temporal basis. This allows for the direct use of the Kronecker product representation without needing to invert the target information matrix first. 

\subsubsection{Regularization}


It should be noted that thus far we have imposed no notion of spatial stationarity or slowly varying characteristics in the model. In behavior learning, it is frequently desirable due to the paucity of samples to incorporate information from adjacent areas when learning the covariance. To achieve this type of gain using the multiresolution model, we obtain additional samples by using slightly shifted copies of the original samples. 

\section{Anomaly Detection Approach}
\label{S:Anom}
\subsection{Model: AR process}
Given the mean and covariance, the standard Mahanalobis distance (Gaussian loglikelihood) is given by Equation \eqref{Eq:Mahanolobis}.

Video is a process, not a single instance. Hence, it is frequently desirable to evaluate the Mahanalobis distance for clips longer than the learned covariance. In order to do this, the larger covariance matrix needs to be inferred from the learned one. A common approach is to assume the process is a multivariate AR process \cite{wiesel2013time}. Time stationary (block Toeplitz where each block corresponds to the correlations between complete frames) covariances define a $T$ length multivariate (in space) AR (in time) process \cite{wiesel2013time}. Using the AR process model, the $T_1> T$ inverse covariance is achieved by block Toeplitz extension \cite{wiesel2013time} of the learned $\Sigma^{-1}$ with zero padded blocks on the $t>T$ super and sub block diagonals. The result is then substituted into Equation \eqref{Eq:Mahanolobis}. A memory efficient implementation is achieved by using
\begin{align}
\label{Eq:AR}
\ell\left(\{x_n\}_{n=1}^{T_1}\right) &= \sum_{n=1}^{T_1}\left((x_n-\mu)^T J_{1} (x_n-\mu) \right)\\\nonumber
&+2\sum_{i=2}^T \sum_{n=1}^{T_1-i+1}(x_n-\mu)^T J_{i} (x_{n+i-1}-\mu)
\end{align}
where $J = \Sigma^{-1}$ and $J_{i} = J_{1:N, (i-1)N+1:iN}$.


\subsection{Anomaly Detection}
Once the likelihood of a video clip has been determined, the result is used to decide whether or not the clip is anomalous.
It is common in anomaly detection to merely threshold the loglikelihood and declare low likelihoods to correspond to anomalies. In the high dimensional regime, however, the distribution of the loglikelihood of an instance given that the instance is generated by the model under which the likelihood is evaluated becomes strongly concentrated about its (nonzero) mean due to concentration of measure. For example, the loglikelihood of a $N$ dimensional Gaussian distribution follows a chi square distribution with $N$ parameters. As a result, high likelihoods are highly unlikely, and are thus probably anomalous. These types of anomalies frequently occur due to excessive reversion to the mean, for example, when everyone in a video leaves the scene. This situation is clearly anomalous (change has occurred) but has a likelihood close to the maximum. Hence, we threshold the likelihood both above and below.


Combination of regions with abnormally high and abnormally low likelihoods can cancel each other out in some cases, resulting in a declaration as normal using the overall likelihood alone. To address this problem, if an instance is determined to be nonanomalous using the overall likelihood, we propose to divide the video into equal sized spatial patches, extract the marginal distributions of each, and compute the loglikelihoods. If the sample variance of these loglikelihoods is abnormally large, then the instance is declared anomalous.

\section{Results}
\label{S:Results}

\subsection{Detection of Crowd Escape Patterns}
\begin{figure}[htb]
\centering
\includegraphics[width=2.85in]{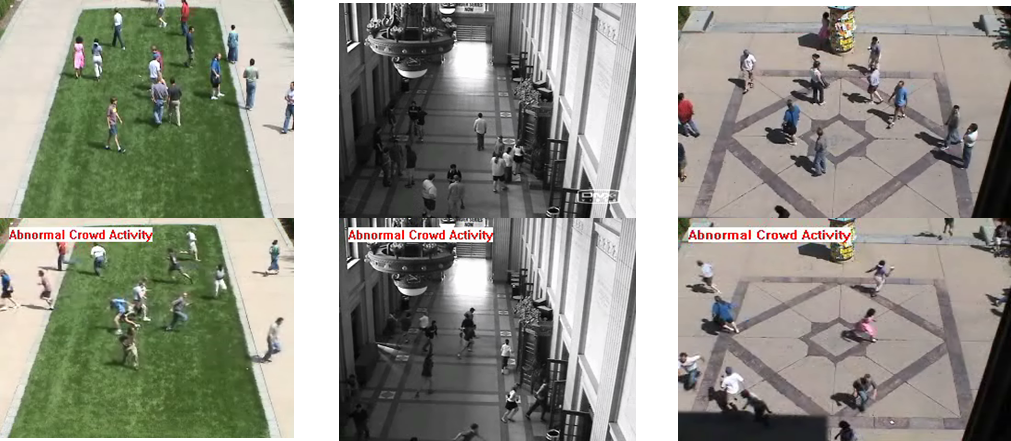}
\caption{Example video frames from each environment in the CMU dataset.}
\label{Fig:Vids}
\end{figure}

To evaluate our methods, we apply them to the University of Minnesota crowd anomaly dataset \cite{UMNData}. This widely used dataset consists of surveillance style videos of normal crowds moving around, suddenly followed by universal escape behavior. There are 11 videos in the dataset, collected in three different environments. Example normal and abnormal frames for each environment are shown in Figure \ref{Fig:Vids}. Our goal is to learn the model of normal behavior, and then identify the anomalous behavior (escape). Since our focus is on anomaly detection rather than identifying the exact time of the onset of anomalous behavior we consider short video clips independently.

\begin{figure}[htb]
\centering
\includegraphics[width=2.85in]{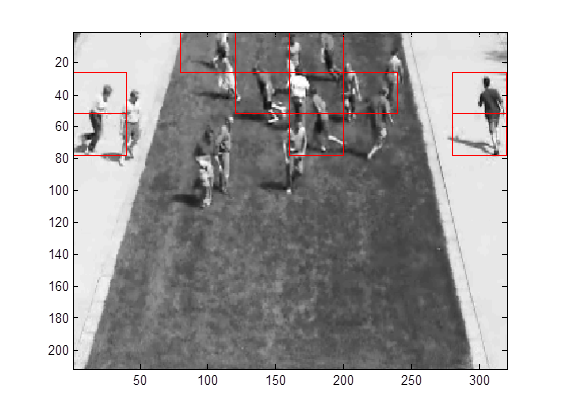}
\caption{Example individual detection results. Blocks declared anomalous are indicated by red boxes. The anomalous behavior is just beginning. Notice the marking of running individuals as anomalous while avoiding the walking individuals.}
\label{Fig:Local}
\end{figure}

Our experimental approach is divide each video into short clips (20-30 frames) and for each clip, use the rest of the video (with the exclusion of a buffer region surrounding the test clip estimate) to estimate the normal space-time pixel covariance. Since the learning the model of normality is unsupervised, the training set always includes both normal and abnormal examples. In essence, then, we are taking each video by itself and trying to find which parts of it are least like the ``average." For simplicity, we convert the imagery to grayscale. The original videos have a label that appears when the data becomes anomalous. We remove this by cutting a bar off the top of the video (see Figure \ref{Fig:Local}). The likelihood of the test clip is then evaluated using the Mahanalobis distance based on the learned spatio-temporal covariance extended into an AR process as in \eqref{Eq:AR}.

Since anomalous regions are included in the training data, the learning of normal behavior is dependent on the preponderance of normal training data, which is the case to some degree. Anomaly detection ROC curves are obtained by optimizing the above and below thresholds following a Neyman-Pearson approach.

\begin{figure}[htb]
\centering
\includegraphics[width=1.85in]{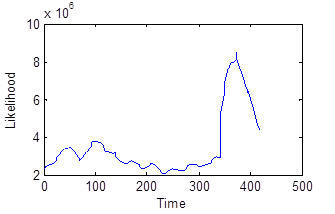}
\caption{Example negative loglikelihood profile (first video). Anomalous behavior begins at the large jump in likelihood. The subsequent decrease is due to people leaving the scene.}
\label{Fig:Likes}
\end{figure}

\begin{figure}[htb]
\centering
\includegraphics[width=2.85in]{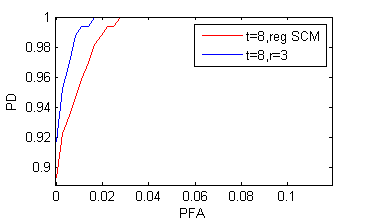}
\caption{ROC curve for anomaly detection on the entire dataset for 8 frame covariance. Thresholds are set for each video individually. The blue curve corresponds to using 3 term sum of Kroneckers (AUC .9995), and the red to the sample covariance with regularization and Toeplitz (AUC .9989). Note the superiority of the Kronecker methods.}
\label{Fig:ROCIndiv}
\end{figure}

\begin{figure}[htb]
\centering
\includegraphics[width=2.85in]{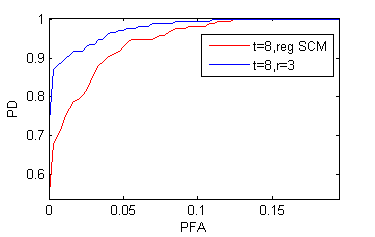}
\caption{ROC curves for anomaly detection on the entire dataset for 8 frame covariance. Thresholds are set for each environment (set of videos) individually. The blue curve corresponds to using 3 term sum of Kroneckers (AUC .995), and the red to the sample covariance with regularization and Toeplitz (AUC .988). Note the superiority of the Kronecker methods.}
\label{Fig:ROCGroup}
\end{figure}

In our first experiment, we use an 8 frame covariance and compare anomaly detection results for the 3 term regularized Toeplitz sum of Kronecker products and the regularized sample covariance with Toeplitz constraint (in other words an 8 term sum of Kronecker products). For mean and covariance estimation, we divide the video into 64 spatial blocks and learn the covariance for each. The test samples are obtained by extracting 30 frame sequences using a sliding window incremented by one frame. The covariance is forced to be the same over sets of 4 blocks in order to obtain more learning examples. The negative loglikelihood profile of the first video as a function of time (frame) is shown in Figure \ref{Fig:Likes} using the Kronecker approach. Note the significant jump in negative loglikilihood when the anomalous behavior begins. Figure \ref{Fig:ROCIndiv} shows the ROC curves for the entire dataset for the case that the thresholds are allowed to be different on each video. Figure \ref{Fig:ROCGroup} shows the results for the case that the thresholds are forced to be constant over the videos in the same environment. This reduces the performance as expected due to less overfitting. Notice that in both cases the use of the Kronecker product representation significantly improves the performance, and that the false alarm rates are quite low.

\begin{figure}[htb]
\centering
\includegraphics[width=2.85in]{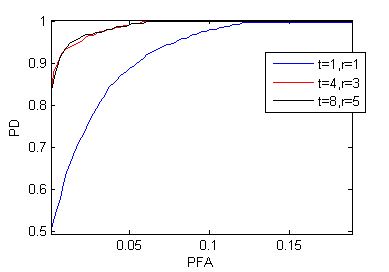}
\caption{ROC curves for 1, 4, and 8 frame covariances. Thresholds are set for each video individually. Note the superiority of multiframe covariance to single frame covariance due to the use of temporal information. }
\label{Fig:TIndiv}
\end{figure}

\begin{figure}[htb]
\centering
\includegraphics[width=2.85in]{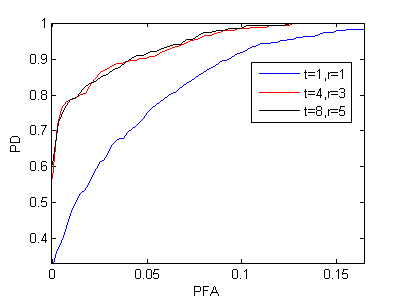}
\caption{ROC curves for 1, 4, and 8 frame covariances. Thresholds are set for each environment (set of videos) individually. Note the superiority of multiframe covariance to single frame covariance due to the use of temporal information.}
\label{Fig:TGroup}
\end{figure}

We then examined the variation of performance with the frame length of the covariance. In this experiment, 20 frame test clips were used and the covariances was held constant over all 64 blocks. Results for 1, 4, and 8 frames are shown in Figures \ref{Fig:TIndiv} and \ref{Fig:TGroup} for individual video and environment thresholding respectively. Note the major gains achieved by incorporating multiframe information.

We also considered localization of the anomalies. This was accomplished by dividing the video into pixel blocks and evaluating the spatio-temporal likelihood of each. Then simple thresholding is used to determine whether or not the patch is anomalous. The results are shown in Figure \ref{Fig:Local}. Note the successful detection of the individuals and only those individuals who have begun running.


\subsection{Detection of Anomalous Patterns in Marathon Videos}
\begin{figure}[htb]
\centering
\includegraphics[width=2.85in]{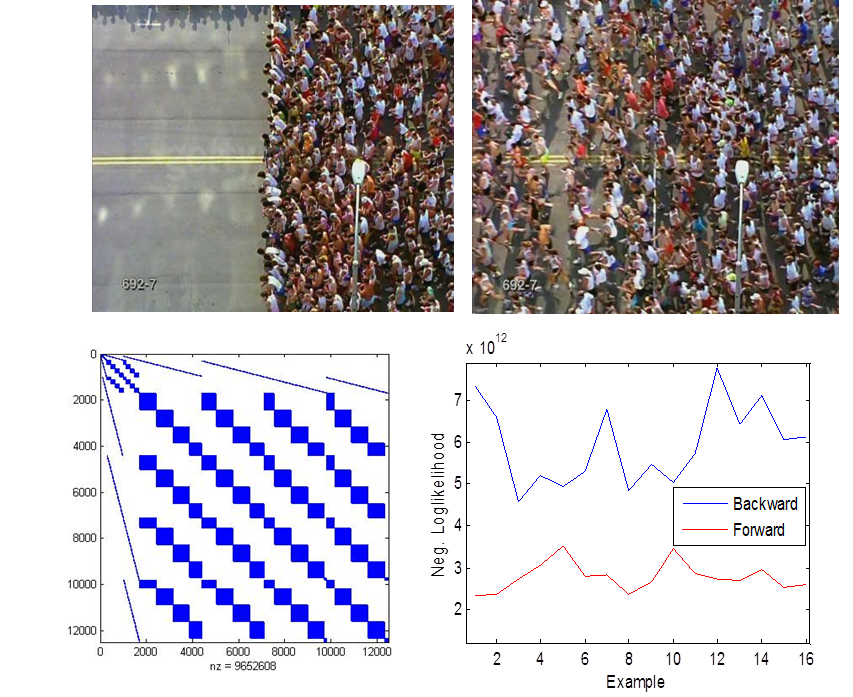}
\caption{Marathon video results using multiresolution model. Upper left: Frame before marathon starts. Upper right: Frame after steady flow has been established. Lower left: Locations of nonzero entries of multiresolution information matrix. Lower right: Negative loglikelihoods as a function of time for test clips from the video and from the video played backwards. }
\label{Fig:Marathon}
\end{figure}

As an example of crowd videos with locally steady optical flow, we consider a video of the start of a marathon, and apply our multiresolution model to learning its covariance. Since steady flow is present, we use nonrectangular tree grids. It was found this was necessary for low separation rank structure to emerge. The model was trained using the same leave out and buffer approach in the previous section. Considering only the portion of the video after the start, we are able to easily determine that clips from the original video are not anomalous whereas the same clips played backwards are anomalous (Figure \ref{Fig:Marathon}).

\section{Conclusion}
\label{Conclusion}
We considered the use of spatio-temporal mean and covariance learning to reliable statistical behavior anomaly detection in video. A major issue with spatio-temporal pixel covariance learning is the large number of variables, which makes sample paucity a severe issue. We found that the approximate pixel covariance can be learned using relatively few training samples using several prior covariance models.

It was found that the space-time pixel covariance for crowd videos can be effectively represented as a sum of Kronecker products using only a few factors, when adjustment is made for steady flow if present. This reduces the number of samples required for learning significantly.

We also used a modified multiresolution model based on \cite{Choi2010gaussian} and incorporating Kronecker decompositions and regularization to decrease the number of required samples to a level that made it possible to estimate the spatio-temporal covariance of the entire image. The learning algorithm in \cite{Choi2010gaussian} was modified to enable significantly more efficient learning.

Using the blockwise Kronecker covariance for the University of Minnesota crowd anomaly dataset, it was found that state of the art anomaly detection performance was possible, and the use of temporal modeling and the sums of Kroneckers representation enabled significantly improved performance. In addition, good anomaly localization ability was observed.

\section{Acknowledgements}
This research was partially supported by ARO under grant W911NF-11-1-0391 and by AFRL under grant FA8650-07-D-1220-0006.

\bibliographystyle{IEEETran}
\bibliography{MultiResKron}

\end{document}